\documentclass[letterpaper, 10 pt, conference]{ieeeconf}  

\IEEEoverridecommandlockouts                              

\usepackage[backref]{hyperref} 
\usepackage[T1]{fontenc}
\usepackage{lmodern}
\usepackage{marvosym}

\hypersetup{hidelinks} 

\usepackage{graphics} 
\usepackage{epsfig} 
\usepackage{mathptmx} 
\usepackage{times} 
\usepackage{amsmath} 
\usepackage{amssymb}  
\usepackage{multirow}
\usepackage{booktabs}
\usepackage{graphicx}
\usepackage{subcaption}
\usepackage{xcolor}
\newcommand{\new}[1]{\textcolor{black}{#1}}
\newcommand{\yuxuan}[1]{\textcolor{black}{#1}}

\title{\LARGE \bf
SCANet: Correcting LEGO Assembly Errors with Self-Correct Assembly Network
}

\author{Yuxuan Wan$^{1,4}$* ~~  Kaichen Zhou$^{2,4}$* ~~  Jinhong Chen$^{3,4}$  ~~ Hao Dong$^{3,4}\dagger$ \\$^{1}$School of Computer Science and Engineering, Southeast University \\ $^{2}$Department of Computer Science, University of Oxford \\ $^{3}$CFCS, School of CS, PKU  \\ $^{4}$PKU-Agibot Lab
\thanks{* Equal contribution (listed alphabetically by last name).
}
\thanks{$\dagger$ Corresponding author.}
}

\begin{document}

\maketitle
\thispagestyle{empty}
\pagestyle{empty}

\begin{abstract}

Autonomous assembly in robotics and 3D vision presents significant challenges, particularly in ensuring assembly correctness. 
Presently, predominant methods such as MEPNet focus on assembling components based on manually provided images. 
However, these approaches often fall short in achieving satisfactory results for tasks requiring long-term planning. Concurrently, we observe that integrating a self-correction module can partially alleviate such issues.
Motivated by this concern, 
we introduce the \textit{ Single-Step Assembly Error Correction Task}, which involves identifying and rectifying misassembled components. 
To support research in this area, we present the LEGO Error Correction Assembly Dataset (LEGO-ECA), comprising manual images for assembly steps and instances of assembly failures. Additionally, we propose the Self-Correct Assembly Network (SCANet), a novel method to address this task. SCANet treats assembled components as queries, determining their correctness in manual images and providing corrections when necessary. Finally, we utilize SCANet to correct the assembly results of MEPNet. 
Experimental results demonstrate that SCANet can identify and correct MEPNet's misassembled results, significantly improving the correctness of assembly. Our code and dataset could be found at \href{https://scanet-iros.github.io/}{https://scanet-iros.github.io/}.
\end{abstract}
\section{Introduction}
\begin{figure}[ht!]
    \centering
    \includegraphics[trim=0cm 4.4cm 14cm 0.55cm, clip, width=\linewidth]{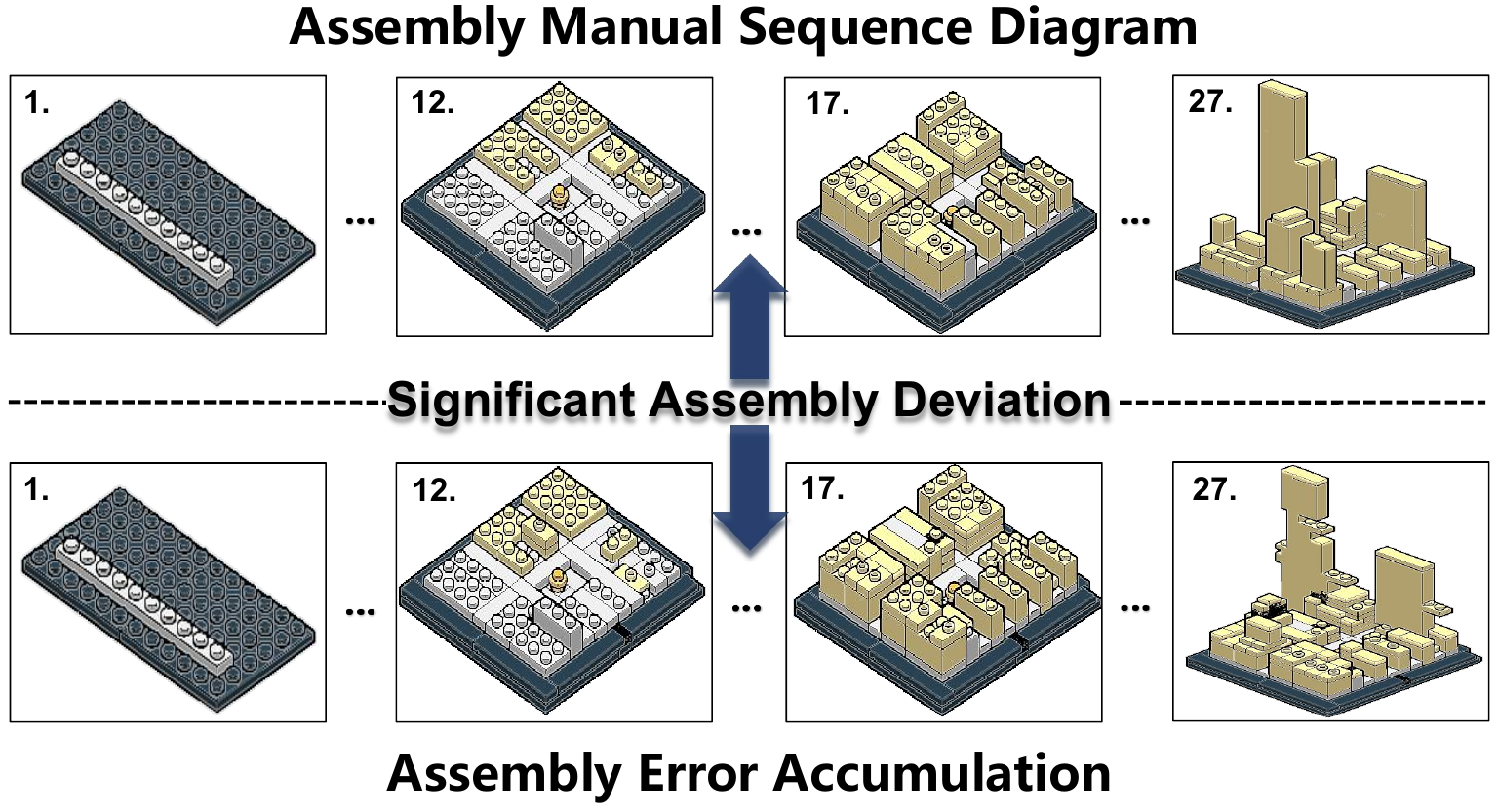}
    \caption{We observed that as the assembly process progresses, assembly errors accumulate, leading to larger discrepancies between the final assembly result and the assembly manual.}
    \label{fig:1}
\end{figure}

For a considerable period, the endeavor to enable robots for autonomous assembly of parts based on human-provided assembly manuals, such as block worlds~\cite{DBLP:conf/iccv/ChenSQ0MZGXXCST19}, LEGO models~\cite{DBLP:conf/nips/ChungKKLTPC21,DBLP:conf/eccv/WangZMCW22}, and furniture~\cite{DBLP:conf/nips/WangZMZCW22,DBLP:journals/scirobotics/ZhouP18}, has been a longstanding pursuit. However, despite the apparent simplicity of such tasks for humans, they pose formidable challenges for robots. Achieving this goal demands extensive engineering efforts and theoretical breakthroughs. Moreover, while humans are prone to errors during part assembly, potentially misplacing various components, they possess the capability to detect and rectify these errors during subsequent assembly processes. Nevertheless, existing research works~\cite{DBLP:conf/eccv/WangZMCW22,DBLP:conf/nips/ZhanFMSCGD20,DBLP:conf/eccv/LiMSSG20,wang2023feature,wu2023leveraging,ogun20153d} have largely overlooked this aspect, expecting robots to execute flawless assembly sequences from start to finish in a single attempt. This expectation proves unrealistic and overly stringent for robots. As assembly progresses, current methods tend to accumulate an increasing number of assembly errors, deviating from the intended sequence. These errors lead to larger discrepancies between the final assembly result and the assembly manual, as illustrated in Figure~\ref{fig:1}. To address this critical research gap, we propose a new task, namely the \textit{``Single-Step Assembly Error Correction Task"}.

\begin{figure*}[tb]
    \centering
    \includegraphics[trim=0.2cm 8.5cm 12.2cm 0.2cm, clip,  width=0.9\linewidth]{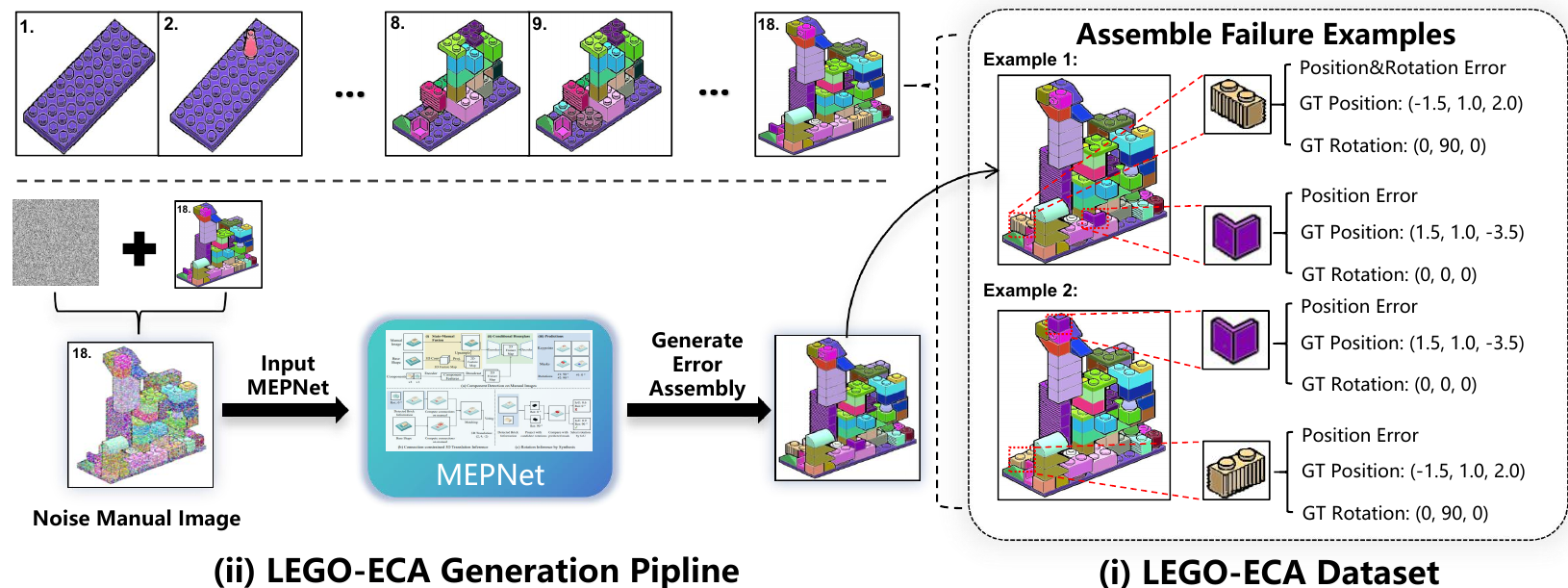}
    \caption{The LEGO-ECA dataset, designed specifically for the single-step assembly error correction task, is presented. (i) Exemplars illustrate assembly manual sequence diagram, instances of component assembly errors, error types, and correct poses. (ii) The construction process outlines how erroneous assembly examples were generated in the LEGO-ECA dataset.}
    \label{fig:2}
\end{figure*}

This task involves identifying components with 6D pose errors in the current assembly step and correcting them. This task presents two new challenges. Firstly, \textit{how to identify misassembled components in the current assembly result?} This requires the ability to discern the current pose of each assembled component, forming the foundational step for conducting assembly error correction. Focusing solely on misassembled components helps mitigate the risk of adjusting correctly assembled components into incorrect pose, thereby reducing the Misplacement Rate (MPR, defined in Section \ref{metrics}). Secondly, \textit{how to correct identified misassembled components?} This involves establishing the relationship between the misassembled component and its correct pose as described in the assembly manual. By doing so, the misassembled component can be adjusted to its correct pose, thus rectifying the assembly error.

To address these challenges, we first constructed a dataset specifically tailored for the single-step assembly correction task, called the LEGO Error Correction Assembly Dataset (LEGO-ECA), which is based on the Synthetic LEGO Dataset~\cite{DBLP:conf/eccv/WangZMCW22}. This dataset comprises 1,429 LEGO assembly manuals, each containing 2D illustrations of each assembly step along with various possible assembly failure scenarios. Further details about the dataset are provided in Section \ref{LEGO-ECA}.

Continuing on, addressing the assembly error correction problem, a natural approach is to compare existing assembly results with assembly manuals, identifying differences to determine which components are misassembled. This approach mirrors human error correction methods. Therefore, we transformed the challenging correction problem into a relatively straightforward query task.
Building upon this idea, we introduce a novel architecture called the \textbf{S}elf-\textbf{C}orrect \textbf{A}ssembly \textbf{Net}work \textbf{(SCANet)}. The core concept of this architecture is to treat each assembled component as a query, querying its assembly status in the manual image, and providing the correct pose when errors occur. More details of SCANet are provided in Section \ref{SCANet}.


To the best of our knowledge, this represents the first attempt to explore and address the single-step assembly error correction task. We constructed the LEGO-ECA dataset and proposed the SCANet architecture to address this task. Ultimately, we utilize the SCANet to correct the assembly results of MEPNet~\cite{DBLP:conf/eccv/WangZMCW22}, and experimental results demonstrate that SCANet can identify and correct MEPNet's misassembled component errors, significantly improving component assembly accuracy.

\section{Related Work}

\subsection{Part Assembly Datasets}
\begin{figure*}[ht!]
    \centering
    \begin{subfigure}[b]{0.32\textwidth}
        \includegraphics[width=\textwidth]{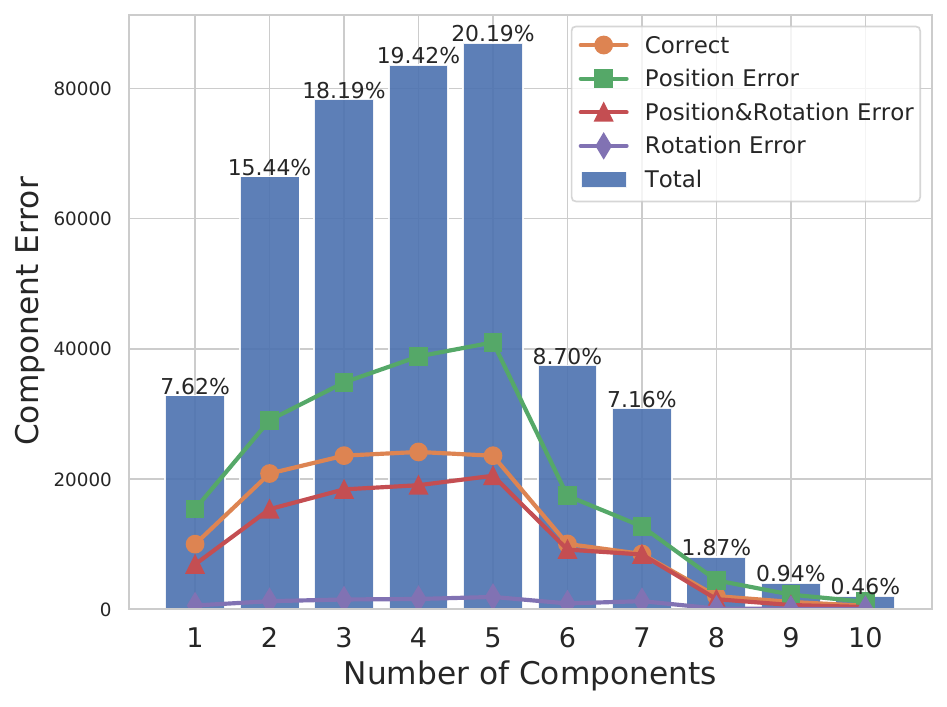}
        \captionsetup{font=footnotesize}
        \caption{Error Components in Single-Step Assembly}
        \label{fig:3a}
    \end{subfigure}
    \begin{subfigure}[b]{0.32\textwidth}
        \includegraphics[width=\textwidth]{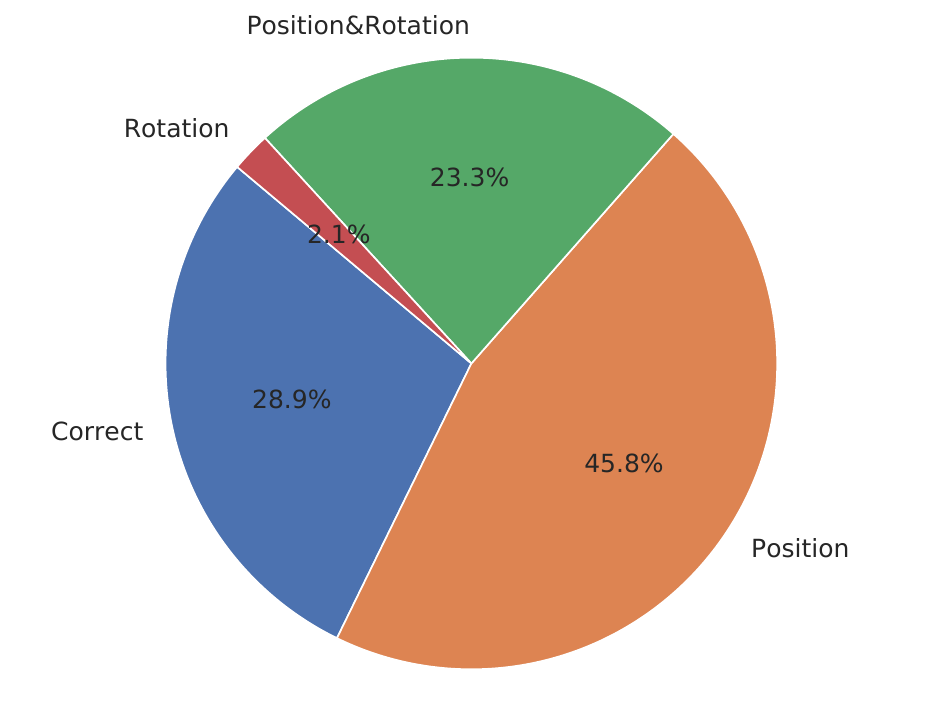}
        \captionsetup{font=footnotesize}
        \caption{Distribution of Different Error Components}
        \label{fig:3b}
    \end{subfigure}
    \begin{subfigure}[b]{0.32\textwidth}
        \includegraphics[width=\textwidth]{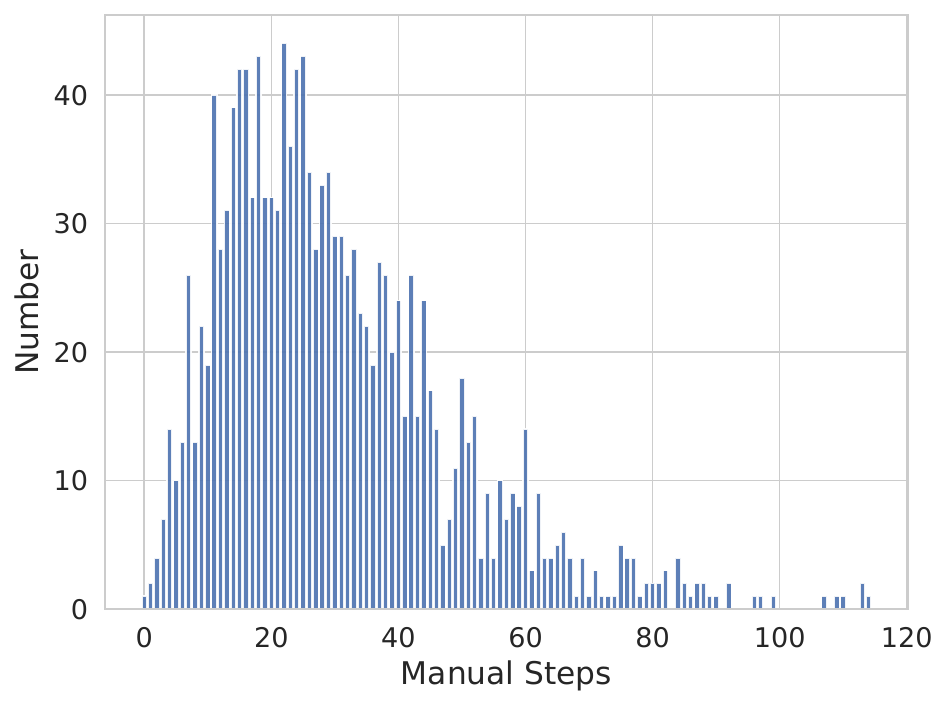}
        \captionsetup{font=footnotesize}
        \caption{\yuxuan{Manuals with different Numbers of Steps}}
        \label{fig:3c}
    \end{subfigure}
    \caption{LEGO-ECA Dataset Statistics. (a) Proportions of different types of incorrectly assembled components in single-step assembly.     In single-step assembly, involving typically 2 to 5 components, positional errors are the most prevalent. Rotational errors frequently accompany positional misalignment and tend to occur concurrently with positional errors, resulting in fewer instances of isolated rotational errors.
    (b) Distribution of different types of incorrectly assembled components across the entire dataset. Positional errors are predominant, while rotational errors are the least frequent. Correct components and those with both positional and rotational errors roughly occupy similar proportions. (c) Statistics of the number of manuals with different numbers of steps. The majority of manuals have step counts ranging from 15 to 40. 
    }
    \label{fig:3}
    \vspace{-0.7cm}
\end{figure*}

In recent decades, numerous 3D part datasets have emerged. Previous works have primarily focused on part assembly \cite{shen2012structure,xie2013sketch,mo2019partnet,zheng2024ha,jones2021automate}. 
The PartNet dataset \cite{mo2019partnet} consists of 573,585 part instances across 26,671 3D models covering 24 object categories. This dataset has facilitated various tasks such as shape analysis, dynamic 3D scene modeling and simulation, affordance analysis, and more. 
Wang \textit{et al.}~\cite{DBLP:conf/nips/WangZMZCW22} introduced IKEA-Manual, a dataset with annotated IKEA objects and assembly manuals. It comprises realistic 3D assembly objects paired with human-designed manuals, parsed in a structured format. Furthermore, the Synthetic LEGO dataset \cite{DBLP:conf/eccv/WangZMCW22} comprises 9,000 LEGO manuals, each presenting a step-by-step instruction sequence for adding new components to an existing LEGO shape. Despite the advancements in prior works, existing datasets have overlooked inherent errors in complex assembly processes. 
In this paper, we aim to bridge this gap by introducing the LEGO Error Correction Assembly (LEGO-ECA) Dataset, which serves as a dataset for assembly error correction task. It is constructed based on the Synthetic LEGO Dataset utilized in MEPNet~\cite{DBLP:conf/eccv/WangZMCW22}. 

\subsection{3D Part Assembly}
Recent advancements \cite{DBLP:conf/eccv/WangZMCW22,DBLP:conf/nips/ZhanFMSCGD20,DBLP:conf/eccv/LiMSSG20,wang2023feature,wu2023leveraging,DBLP:conf/cvpr/ChenLTJG22,DBLP:conf/iclr/LiuZHH0Y23,DBLP:conf/wacv/HarishNR22,DBLP:conf/icra/LeeHL21,DBLP:journals/corr/abs-2311-16592} in assembly guidance leverage cutting-edge technologies to improve efficiency and accuracy. 
Huang \textit{et al.}~\cite{DBLP:conf/nips/ZhanFMSCGD20} pioneers a dynamic graph learning framework, predicting part poses within point clouds by reasoning over their relations using dynamic modules.  
Li \textit{et al.}~\cite{DBLP:conf/eccv/LiMSSG20} proposes a two-module pipeline for single-image-guided 3D part assembly, emphasizing accurate pose prediction and relationship constraints through 2D-3D correspondences and graph message-passing. Additionally,  Wang \textit{et al.}~\cite{DBLP:conf/eccv/WangZMCW22} introduces MEPNet, employing neural keypoint detection and 2D-3D projection algorithms for reconstructing assembly steps from manual images. 
These innovative approaches signify a paradigm shift in assembly processes, promising enhanced precision and automation. 
While the mentioned methods~\cite{DBLP:conf/eccv/LiMSSG20,DBLP:conf/nips/ZhanFMSCGD20,DBLP:conf/eccv/WangZMCW22} focus primarily on how to assemble components, they overlook the potential for errors in this process. Our work takes a pivotal turn by not only addressing the act of assembly but also delving deeper into the crucial aspect of ensuring its accuracy. We go beyond mere assembly guidance to incorporate mechanisms for error detection and correction.

Thus, our approach marks a significant advancement, shifting the emphasis from solely executing assembly tasks to actively verifying and rectifying any inaccuracies in the assembled components.

\section{LEGO Error Correction Assembly Dataset}




\label{LEGO-ECA}

The LEGO Error Correction Assembly Dataset (LEGO-ECA) is a dataset constructed for the single-step assembly correction task, based on the Synthetic LEGO dataset used in MEPNet~\cite{DBLP:conf/eccv/WangZMCW22}. The construction process and composition of the dataset are illustrated in Figure \ref{fig:2}. It comprises 1,429 LEGO assembly manuals, each containing 2D illustrations of each assembly step and various possible assembly failure examples. For each assembly failure example, as  illustrated in Figure \ref{fig:2}(i), we provide a 2D rendering of the failed assembly and the corresponding 3D shape. Additionally, we annotate the error types of each incorrectly assembled component and the correct assembly pose. 

To construct the LEGO-ECA dataset, as shown in Figure \ref{fig:2}(ii), we randomly selected 1,429 assembly manuals from the training set of the Synthetic LEGO Dataset. Subsequently, we inputted each step into MEPNet for component assembly. 
To simulate incorrect assembly samples more effectively and consistently, we preprocessed the manual images before inputting them into MEPNet. This involved adding random Gaussian noise to interfere with the assembly process, thereby inducing MEPNet to output incorrectly assembled examples.
For each assembly step, we applied 3 to 5 random Gaussian noise injections. Ultimately, we constructed a dataset containing approximately 120,000 instances of incorrectly assembled examples. We conducted statistical analysis on the LEGO-ECA dataset, and the relevant results are shown in Figure \ref{fig:3}. Figure \ref{fig:3a} depicts the ratio of components in different states to the total number of components used in single-step assembly. Figure \ref{fig:3b} presents the ratio of different error types of components to all components. 
Figure \ref{fig:3c} illustrates the distribution of the number of manuals with different numbers of steps.


\section{Self-Correct Assembly Network}
\label{SCANet}

\begin{figure*}
    \centering
    \includegraphics[trim=0cm 5.8cm 1cm 0, clip, width=\linewidth]{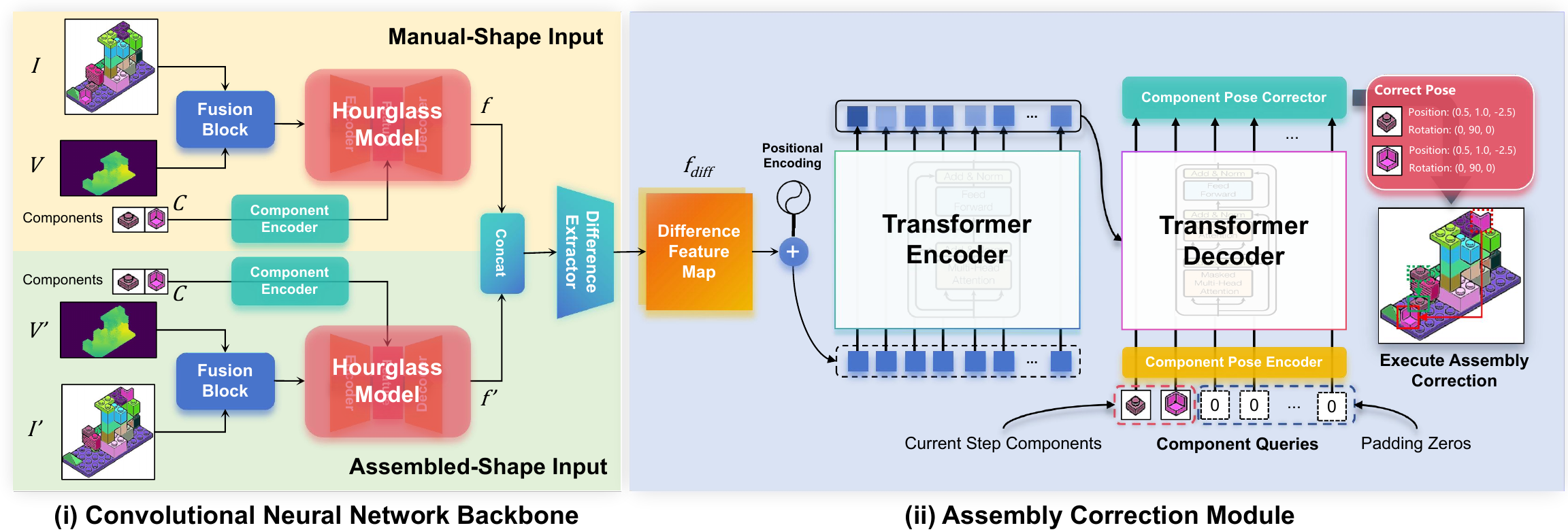}
    \caption{SCANet consists of two modules. (i) a convolutional neural network backbone, comprising a fusion block, Hourglass model, and assembly difference extractor, which extracts differential features between manual images and assembly results; and (ii) an assembly correction module, the core of SCANet, consisting of three parts: a component pose encoder, transformer network, and component pose corrector, which outputs the final corrected component pose information.}
    \label{fig:4}
    \vspace{-0.3cm}
\end{figure*}


\subsection{Problem Formulation}
In this context, we define the \textit{Single-Step Assembly Correction Task}. Firstly, the prerequisite for performing the assembly correction task is completing the component assembly task. Therefore, we denote all prerequisites needed for the assembly task as \( C_a \), which include 2D manual images \( I \), the initial shapes before assembly \( V \), and all components used \( C = \{c_1, c_2, \ldots, c_N\} \). Additionally, assembly correction requires the component poses after assembly \( P_i = \{(R_i, T_i) | R_i \in \mathbb{R}^{3 \times 3}, T_i \in \mathbb{R}^{3}, i = 1,2, \ldots, N\} \) obtained by the model \( M \), the resulting shape after assembly \( V' \), and  its corresponding 2D rendered image \( I' \). The objective of our assembly correction task is as follows: based on \( C_a \), given \( V' \) and \( I' \) obtained after assembly by the assembly model \( M \), predict the status \( S_i \) of each component's pose \( P_i \), and correct each \( S_i \) predicted as an erroneous component pose \( P_i^{error} \) to the correct component pose \( P_i^{correct} \).

\subsection{Architecture Overview}

Referencing Figure \ref{fig:4}, the architecture of our proposed model, SCANet, shares similarities with DETR~\cite{carion2020end} and is broadly divided into two main modules. The first module, shown in Figure \ref{fig:4}(i), is a convolutional neural network backbone that plays a crucial role in extracting the difference feature \( f_{\text{diff}} \) between the assembly manual image and the assembly result. It takes as inputs the images \( I \) and \( I' \), the 3D shapes \( V \) and \( V' \), and the components \( C \). The second module, depicted in Figure \ref{fig:4}(ii), is the assembly correction module, which is responsible for determining the correctness of the component assembly and correcting any misassembled parts. This module processes the features \( f_{\text{diff}} \) produced by the backbone, the components \( C_{i} \) used in the current step, the current pose \( P_{i} \), and the 2D image \( I'_C \) of each component. It then outputs the status \( S_{i} \) of each component, indicating whether it has been correctly assembled, along with the 6D pose information \( P'_{i} \) for each component after correction.

\subsection{Convolutional Neural Network Backbone}

As illustrated in Figure \ref{fig:4}(i), the convolutional neural network backbone of SCANet shares a similar structure to MEPNet~\cite{DBLP:conf/eccv/WangZMCW22}, incorporating the image-shape fusion block and the Hourglass model~\cite{newell2016stacked}. However, SCANet diverges from MEPNet with its inclusion of two input branches: the Manual-Shape branch and the Assemble-Shape branch. These branches share identical network structures and weight parameters. Additionally, SCANet integrates an assembly difference extractor to capture the disparity between the assembly result and the assembly manual. 
Specifically, the image \(I\), 3D shape \(V\), and components \(C\) are fed into the Manual-Shape branch, yielding feature \(f\) after passing through the fusion block and Hourglass model. Similarly, the image \(I'\), 3D shape \(V'\), and components \(C\) are input to the Assemble-Shape branch, generating feature \(f'\). Both features possess dimensions of \(\mathbb{R}^{C_{1} \times H_1 \times W_1}\), where \(H_1, W_1 = \frac{H_{\text{img}}}{4}, \frac{W_{\text{img}}}{4}\), and \(C_1=128\).
Subsequently, by concatenating features \( f \) and \( f' \) along the channel dimension, a new feature \( f_{cat} \) is formed with dimensions \( \mathbb{R}^{2C_{1} \times H_1 \times W_1} \). Finally, \( f_{cat} \) undergoes processing by the assembly difference extractor to derive the difference feature \( f_{diff} \) with dimensions \( \mathbb{R}^{C_{2} \times H \times W} \), where \( C_2 = 256 \), and \( H, W = \frac{H_{\text{img}}}{16}, \frac{W_{\text{img}}}{16} \). The assembly difference extractor comprises two residual blocks. 

\subsection{Assembly Correction Module}
As depicted in Figure \ref{fig:4}(ii), the assembly correction module of SCANet serves as the core module of the network and comprises three parts: the component pose encoder, transformer network, 
and component pose corrector. The inputs to the entire module include the difference feature \( f_{\text{diff}} \) obtained from the backbone, the 3D voxel data  \( C \) of the components, the 2D image \( I'_C \) of the components after assembly, and the pose information \( P \) of the components. The outputs include the status information \( S \) of the components and the corrected components' pose \( P' \).

\begin{figure}
    \centering
    \includegraphics[trim=0cm 5.5cm 23cm 0, clip, width=0.85\linewidth]{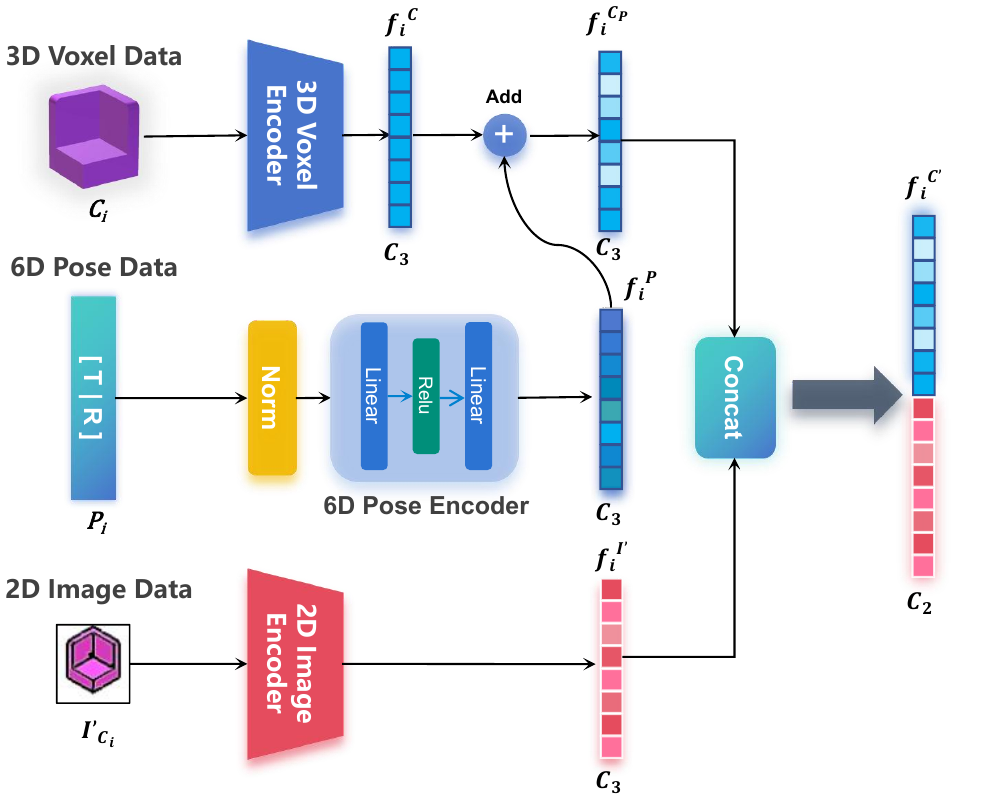}
    \caption{The component pose encoder consists of three sub-encoders: a 3D voxel encoder, a 6D pose encoder, and a 2D image encoder. The 3D voxel encoder encodes the component's 3D voxel data into a one-dimensional vector, which is then combined with the output of the 6D pose encoder. This combined vector is then concatenated with the 2D image encoder's output, producing a component feature that integrates 6D pose, 2D image, and 3D shape information.}
    \label{fig:5}
\end{figure}

\textbf{Component Pose Encoder:} As illustrated in Figure \ref{fig:5}, the component pose encoder comprises three sub-encoders: the 3D voxel encoder, the 6D pose encoder, and the 2D image encoder. Firstly, the 3D voxel encoder adopts a structure similar to the component encoder in MEPNet~\cite{DBLP:conf/eccv/WangZMCW22}. It consists of multiple 3D Convolution-BatchNorm-ReLU layers, followed by an average pooling layer and a flatten layer. This process encodes the 3D voxel data into a one-dimensional vector feature represented as \(f_i^C \in \mathbb{R}^{C_3}\).
Next, the 6D pose encoder comprises two fully connected layers. It takes the component's 6D pose information \(P=[T|R]\), where \(P \in \mathbb{R}^6\). \(T \in \mathbb{R}^3\) represents the component's coordinates in 3D space, and \(R \in \mathbb{R}^3\) denotes the Euler angle representation of the component. The output is the encoded 6D pose feature \(f_i^P \in \mathbb{R}^{C_3}\). Before feeding into the 6D pose encoder, we normalize each component's \(T_i\) and \(R_i\) separately according to the following formula:
\begin{equation} \left\{\begin{array}{l} T^x_i, R^x_i= \frac{T^x_i-T_{min}^{x}}{T^{x}_{max}-T^{x}_{min}}, \frac{R^x_i-R^{x}_{min}}{R^{x}_{max}-R^{x}_{min}}
\\ T^y_i,R^y_i= \frac{T^y_i-T^{y}_{min}}{T^{y}_{max}-T^{y}_{min}},\frac{R^y_i-R^{y}_{min}}{R^{y}_{max}-R^{y}_{min}}
\\ T^z_i,R^z_i= \frac{T^z_i-T^{z}_{min}}{T^{z}_{max}-T^{z}_{min}},\frac{R^z_i-R^{z}_{min}}{R^{z}_{max}-R^{z}_{min}}
\end{array}\right. .\end{equation} 
We then add $f_i^P$ and $f_i^C$ to obtain a component feature $f_i^{C_P} \in \mathbb{R}^{C_3}$, which encodes the 6D pose information. Subsequently, for the 2D image encoder, we employ the ResNet network~\cite{kocabas2018multiposenet} structure, taking the 2D image data \( I'_{C_i} \) of each component after assembly as input, and producing the encoded component image feature \( f_i^{I'} \in \mathbb{R}^{C_3} \). Here, we modify the output dimensionality of the ResNet. Finally, we concatenate \( f_i^{C_P} \) and \( f_i^{I'} \) along the channel dimension, yielding a component feature \( f_i^{C'} \in \mathbb{R}^{C_2} \) that integrates both the 6D pose information and the 2D image information simultaneously.



\begin{figure}
    \centering
    \includegraphics[trim=0cm 10cm 23cm 0, clip, width=0.85\linewidth]{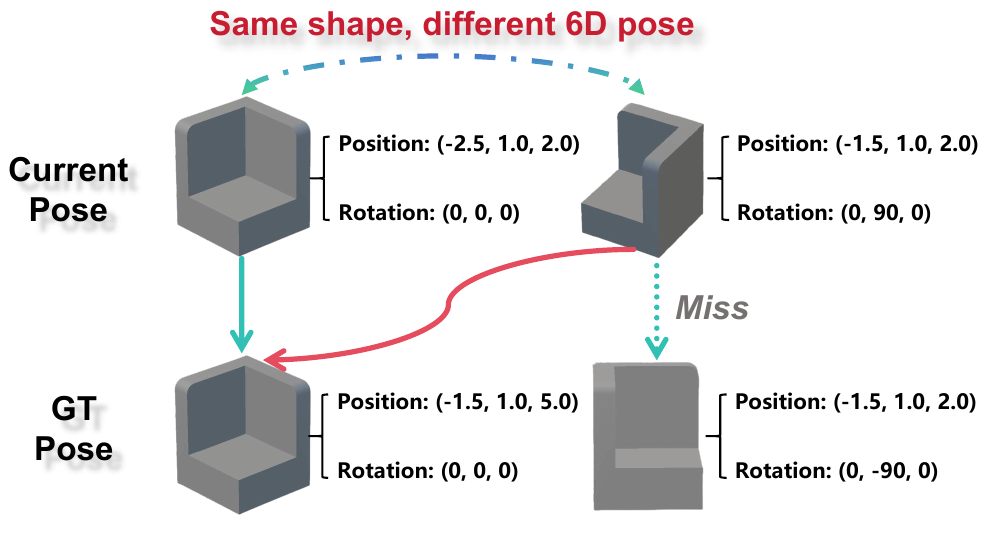}
    \caption{\new{SCANet may misalign multiple components of the same shape to the same GT pose, incorporating color information can help mitigate this issue.}}
    \label{fig:6}
    \vspace{-0.2cm}
\end{figure}

\new{Encoding both 6D pose and 2D image information is essential. During the correction process, SCANet may mistakenly adjust multiple components of the same shape to the same GT pose, as shown in Figure~\ref{fig:6}. Incorporating image data, including color information, helps alleviate this issue. Additionally, 3D voxel data and 2D image data are complementary—relying solely on 2D image data sacrifices geometric detail, while using only 3D voxel data omits the color information of the components.}

\textbf{Transformer Network:} The encoder and decoder of the transformer model in our approach closely resemble those in DETR~\cite{carion2020end}, following the standard transformer architecture. However, there are slight differences in our decoder. Firstly, while DETR uses learned object query embeddings, our component query embeddings are obtained by encoding through the component pose encoder. Secondly, in DETR, the number of object queries input into the decoder is fixed, whereas the length sequence of component queries we input varies. For each batch, the length of queries follows the maximum number of component queries in that batch, with any remaining samples having fewer queries padded with zero vectors.

\begin{figure}
    \centering
    \includegraphics[trim=0cm 5.5cm 28cm 0, clip, width=0.6\linewidth]{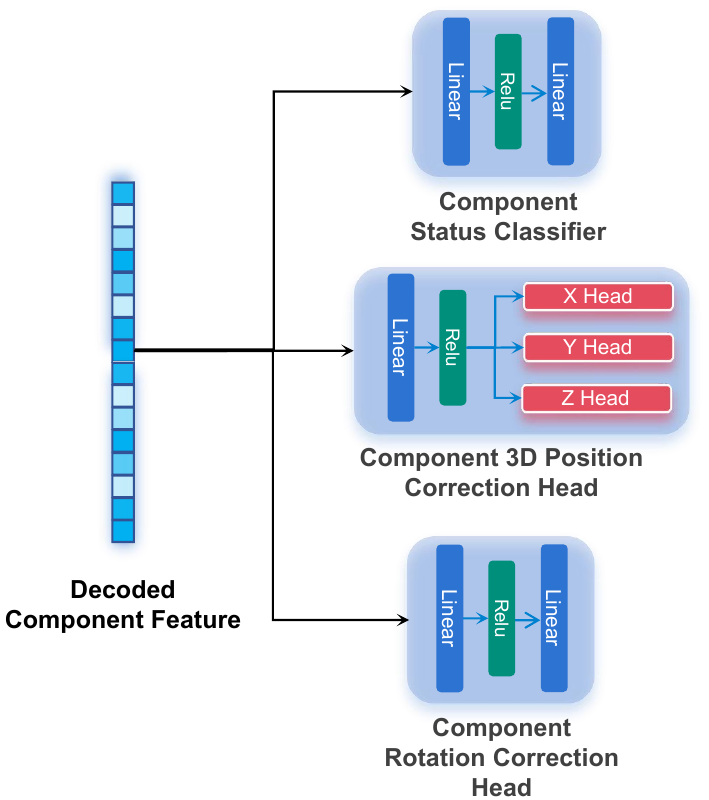}
    \caption{\new{The component pose corrector includes three distinct fully connected neural networks: a status classifier, a 3D position correction head, and a rotation correction head.}}
    \label{fig:7}
\end{figure}
\textbf{Component Pose Corrector:}
As illustrated in Figure~\ref{fig:7}, the component pose corrector comprises three small fully connected neural networks with distinct structures: the component status classifier, the component 3D position correction head, and the component rotation correction head.
\begin{enumerate}
    \item \textbf{Component Status Classifier:} This classifier consists of two fully connected layers. Its output, a feature vector \( f_{1d} \in \mathbb{R}^{C_4} \), where \( C_4=4 \), encodes four possible component statuses: correctly assembled, position error, rotation error, or both position and rotation errors.
    \item \textbf{Component 3D Position Correction Head:} We adopt the structure of SimCC~\cite{li2022simcc}, treating each component's 3D coordinates as a classification task in three directions (x, y, z). Specifically, after processing through this head, it will output a tuple consisting of three 1D feature maps \((f_x, f_y, f_z)\). Additionally, the size of each feature map is consistent with the size of the voxel grid in each direction.
    \item \textbf{Component Rotation Correction Head:} Following a similar approach to MEPNet, we treat different rotation angles (0°, 90°, 180°, 270°) as a classification task. However, we simplify the treatment of rotational symmetry. While MEPNet considers rotational symmetry in the classification task, we do not consider it when outputting the rotation correction result. Instead, symmetry of the component is considered during specific correction. For example, if a component is symmetry at 0° and 180°, its rotation is considered as 0°.
\end{enumerate}

\subsection{Training and Loss}
We train SCANet on the LEGO-ECA dataset, which provides component status, 3D position, and rotation information. The network is trained end-to-end using gradient descent. Our objective function is defined as:

\begin{equation}
L=\alpha \cdot L_{position}+\beta \cdot L_{rotation}+\gamma \cdot L_{status},
\end{equation}
here, \( L_{position} \), \( L_{rotation} \), and \( L_{status} \) are all cross-entropy losses. Unlike MPNet, which employs various loss functions for training, we simplify all tasks into classification tasks, making the overall network easier to train. Additionally, we adopt auxiliary loss settings inspired by DETR~\cite{carion2020end} to further accelerate network convergence. Specifically, we provide the output of each layer of the transformer decoder to the component pose corrector, calculate the loss based on the corrector's output, and then take the mean loss of all layers.
\section{Experiments}

\begin{table*}
\centering
    \begin{tabular}{llcccccc} 
    \toprule
    \multirow{2}{*}{Dataset}       & \multirow{2}{*}{Method} & \multicolumn{2}{c}{Pose Acc $\uparrow$ (\%)}   & \multirow{2}{*}{MTC $\downarrow$ (\%)} & \multirow{2}{*}{CR $\uparrow$ (\%)} & \multirow{2}{*}{MPR $\downarrow$ (\%)} & \multirow{2}{*}{CD $\downarrow$}    \\ 
    \cline{3-4}
                                   &                         & Component        & Step             &                           &                          &                           &                      \\ 
    \hline
    \multirow{3}{*}{LEGO-ECA}  & MEPNet~\cite{DBLP:conf/eccv/WangZMCW22}                  & 38.58         & 28.66          & 66.46                   & --                       & --                        &28.98                      \\
                                  
                                   & SCANet                  & ~~~~~~\textbf{49.31}$^{{\tiny \uparrow 10.73}}$ & ~~~~~\textbf{36.96}$^{{\tiny \uparrow 8.30}}$ & ~~~~~~\textbf{58.07}$^{{\tiny \downarrow 8.33}}$          &  \textbf{ 19.31 }                      &  \textbf{6.35}                       &  \textbf{12.61}$^{{\tiny \downarrow 16.37}}$                     \\ 
    \hline
    \multirow{3}{*}{Synthetic-LEGO} & MEPNet~\cite{DBLP:conf/eccv/WangZMCW22}                  & 33.70          & 24.94          & 69.80                   & --                       & --                        &   40.02                   \\
                                   & SCANet                  & ~~~~~\textbf{38.29}$^{{\tiny \uparrow 5.22}}$ & ~~~~~\textbf{29.24}$^{{\tiny \uparrow 4.30}}$ & ~~~~~\textbf{64.92}$^{{\tiny \downarrow 4.88}}$          & \textbf{9.02}         & \textbf{2.47}           &      \textbf{24.08}$^{{\tiny \downarrow 15.94}}$                \\
    \bottomrule
    \end{tabular}
    \caption{Experimental results of assembly correction on LEGO-ECA and Synthetic-LEGO datasets. \new{Here, MEPNet is used for assembly, while SCANet is applied to correct the results produced by MEPNet.} Chamfer distance (CD) metrics are multiplied by a factor of $10^5$.}
    \vspace{-0.5cm}
\label{tab:table1}
\end{table*}

\begin{table}
\centering
\resizebox{\linewidth}{!}{
    \begin{tabular}{lcccc} 
    \toprule
    \multirow{2}{*}{Method} & \multicolumn{2}{c}{Pose Acc $\uparrow$ (\%)} & \multirow{2}{*}{CR $\uparrow$ (\%)} & \multirow{2}{*}{MPR $\downarrow$ (\%)}  \\ 
    \cline{2-3}
                            & Component & Step                  &                          &                            \\ 
    \hline
    MEPNet~\cite{DBLP:conf/eccv/WangZMCW22}                  & 38.58   & 28.66               & --                       & --                         \\
    \hline
    MEPNet*                 & 42.04   & 32.48               &    14.63                      &      ~9.74                      \\
    SCANet {\tiny (w.o. AR) }    & 48.78   &   36.10            &   19.23                       &  ~8.44                          \\
    SCANet {\tiny (w.o. IM)}    & 45.99  &     33.17          &   \textbf{23.40}                       &   18.58                         \\
    SCANet {\tiny (w.o. 6D\&IM)}    & 35.80    &  28.46               &  14.54                        & 23.25                           \\
    \hline

    SCANet                  & \textbf{49.31}   & \textbf{36.96}               & 19.31                           & \textbf{~6.35  }                            \\
    \bottomrule
    \end{tabular}
}
    \caption{
    Ablation experiments of SCANet structure on LEGO-ECA. `MEPNet*' denotes our adapted MEPNet. `(w.o. AR)' indicates the absence of the assembly result branch in the convolutional neural network backbone. `(w.o. IM)' signifies the removal of the 2D image encoder from the component pose encoder. `(w.o. 6D\&IM)' denotes the removal of both the 2D image encoder of the component pose encoder and the 6D pose encoder.
    }
    \vspace{-0.6cm}

\label{tab:table2}
\end{table}

\subsection{Implementation Details}

SCANet is trained for 100 epochs using the AdamW optimizer with a 1e-4 learning rate, a batch size of 8, and gradient accumulation (step size 4). Empirically, the loss function hyperparameters \( \alpha \) and \( \beta \) are set to 1, while \( \gamma \) is 0.5. The image input size is fixed at \( 512 \times 512 \), consistent with MEPNet~\cite{DBLP:conf/eccv/WangZMCW22}. Both the Transformer~\cite{DBLP:conf/nips/VaswaniSPUJGKP17} and ResNet~\cite{kocabas2018multiposenet} in SCANet utilize pretrained models.

\subsection{Dataset}


We first randomly partition the LEGO-ECA dataset into 80\% for training and 20\% for validation. During both training and validation, we ensure that only components with errors in the current step are considered, meaning that components from non-current steps are correctly assembled. For testing assembly error correction, we randomly select 10\% of the LEGO-ECA dataset (150 manuals). In this phase, we adopt a setwise approach similar to MEPNet~\cite{DBLP:conf/eccv/WangZMCW22}, where we correct the assembly errors of the current step and use the corrected result as input for the next step of assembly. However, since our correction process does not guarantee 100\% accuracy, some incorrectly assembled components may still be passed on to the subsequent step (as illustrated in Figure \ref{fig:6} (d)).

To further assess the generalization capability of our method, we conduct additional experiments by randomly selecting 150 assembly manuals from the Synthetic LEGO dataset for testing. Importantly, these 150 test manuals have no overlap with the manuals in the LEGO-ECA dataset.


\subsection{Evaluation Metrics}
\label{metrics}
To evaluate the performance of SCANet on the single-step assembly error correction task, we employed various metrics. Firstly, we referred to some evaluation metrics from MEPNet~\cite{DBLP:conf/eccv/WangZMCW22}, including the component pose accuracy after correction, \textit{Chamfer distance}~\cite{DBLP:conf/cvpr/FanSG17}, and step-wise pose accuracy (considering a step correct if all component predictions in that step are correct). Secondly, as we adopted the setwise setting from MEPNet, we computed the normalized \textit{Mistakes to Complete (MTC)} score~\cite{DBLP:conf/iccv/ChenSQ0MZGXXCST19}. Lastly, we introduced two additional metrics to assess the correction effectiveness, namely\textit{ Correction Rate} (CR) and \textit{Misplacement Rate} (MPR).

We define the Correction Rate as the ratio of the number of originally incorrect components corrected to the correct pose to the total number of originally incorrect components, formulated as follows:

\begin{equation}
CR=\frac{N_c}{N_{ic}}   , 
\end{equation}
here, \(N_c\) represents the number of originally incorrect components corrected to the correct pose, and \(N_{ic}\) denotes the total number of originally incorrect components.

We define the Misplacement Rate as the ratio of the number of originally correct components corrected to the wrong pose to the total number of originally correct components, expressed as:

\begin{equation}
MPR=\frac{N_w}{N_{c}},
\end{equation}
here, \(N_w\) represents the number of originally correct components corrected to the wrong pose, and \(N_{c}\) indicates the total number of originally correct components.

\subsection{Results and Analysis}

\noindent \textbf{Assembly Error Correction Experiments.} We conducted single-step assembly error correction experiments on the test sets from the LEGO-ECA dataset and a test set extracted from the Synthetic LEGO dataset. As shown in Table \ref{tab:table1}, the overall assembly performance significantly improves when using SCANet to correct the assembly results produced by MEPNet. 
For the LEGO-ECA test set, we observe relatively low assembly accuracy of MEPNet during consecutive assembly steps. However, after correction by SCANet, the accuracy of MEPNet's assembly results improves significantly. This result validates the ability of our SCANet to correct incorrectly assembled components. Regarding the Synthetic-LEGO test set, which neither SCANet nor MEPNet had seen during training, SCANet is still able to correct the assembly results produced by MEPNet. This further demonstrates the generalization capability of our method, indicating its ability to correct errors in completely unseen components.

\begin{figure}[ht!]
    \centering
    \includegraphics[width=0.95\linewidth]{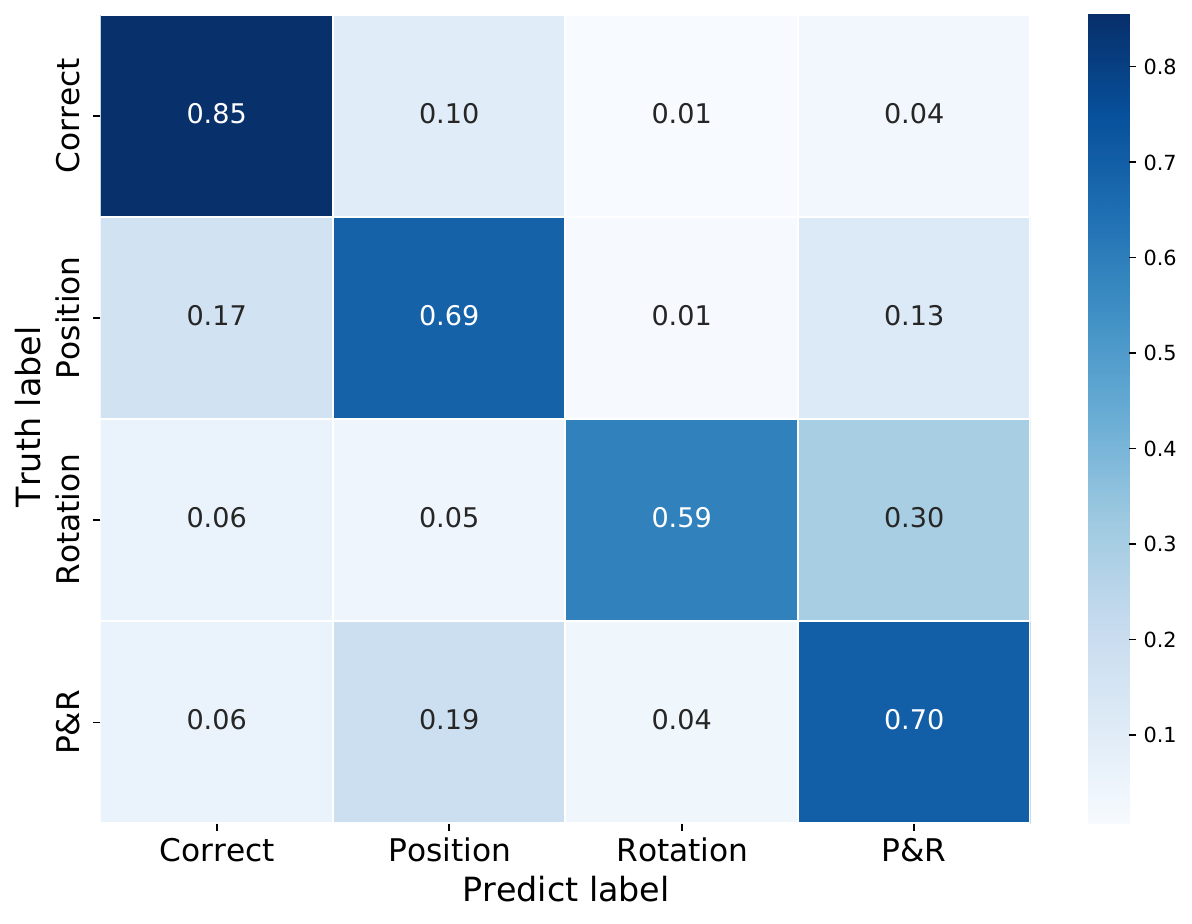}
    \caption{\new{Visualization of the confusion matrix for assembled results. The term `Correct' denotes successful assembly, `Position' refers to errors in assembly position, `Rotation' indicates errors in assembly rotation, and `P\&R' represents errors in both position and rotation.}}
    \label{fig:8}
\end{figure}

\noindent \new{\textbf{Confusion Matrix Visualize Results.} As illustrated in Figure~\ref{fig:8}, we utilized a confusion matrix to visualize SCANet's classification of MEPNet's assembly results. This visualization aids in a deeper understanding of SCANet’s performance in recognizing various assembly errors. The figure reveals that SCANet effectively distinguishes correctly assembled components and demonstrates good performance in identifying position errors, rotation errors, and position\&rotation errors. However, SCANet exhibits a tendency to confuse rotation errors with position\&rotation errors. This confusion arises because rotation errors often cause slight shifts in the component's center of mass (especially for asymmetrical components), leading SCANet to mistakenly interpret these shifts as the position\&rotation errors.}

\begin{figure}[ht!]
    \centering
    \includegraphics[trim=0.3cm 3cm 12.5cm 0.8cm, clip, width=\linewidth]{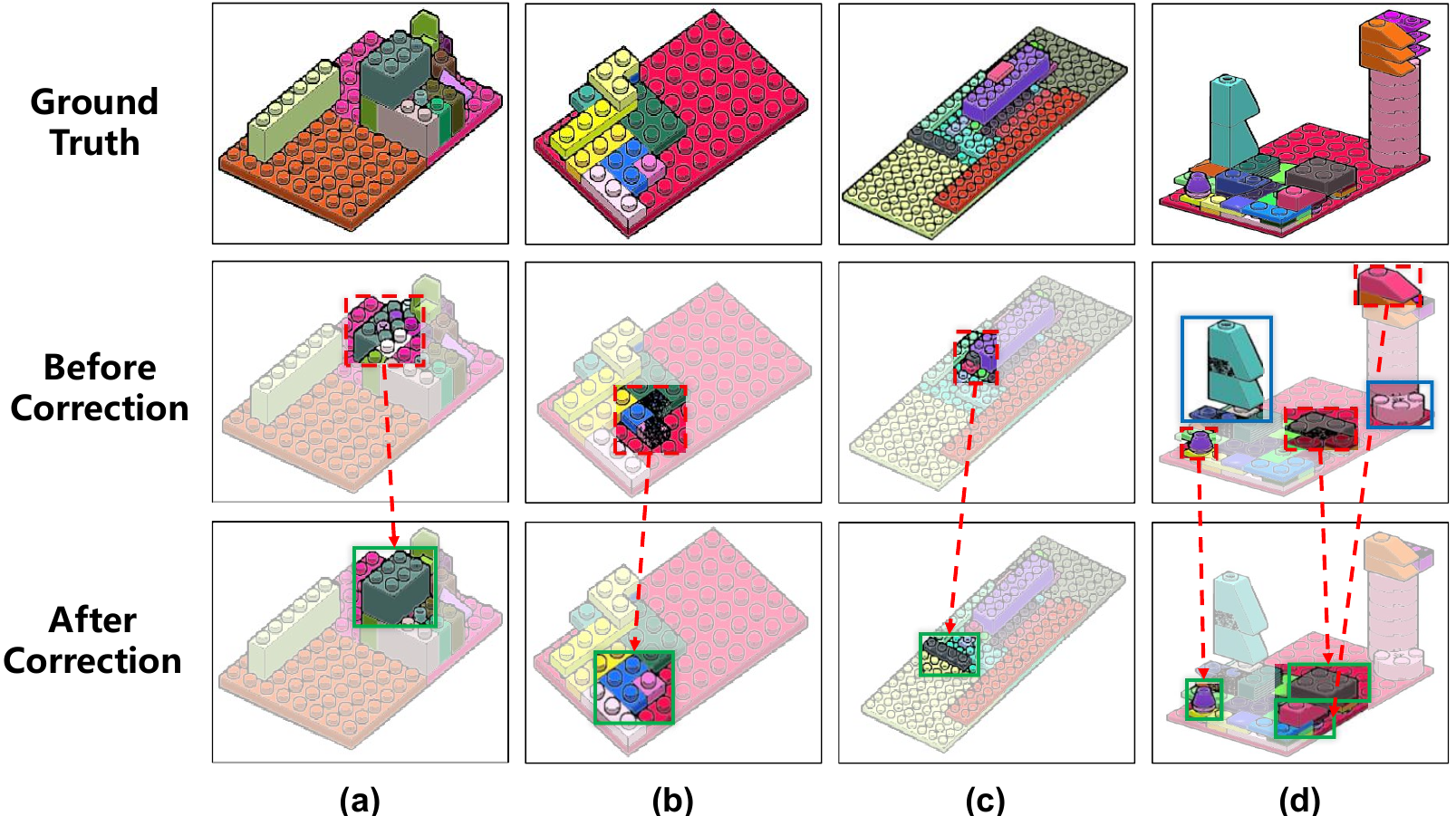}
    \caption{Visualization of SCANet correcting misassembled components. \textbf{\textcolor{red}{Red boxes}} show components corrected before error correction, \textbf{\textcolor{green}{green boxes}} show components corrected after error correction, and \textbf{\textcolor{blue}{blue boxes}} indicate assembly errors that remained uncorrected (\textit{i.e.}, accumulated errors).}
    \label{fig:9}
    \vspace{-0.3cm}
\end{figure}

\noindent \textbf{Assembly Correction Visualize Results.} As shown in Figure \ref{fig:9}, we visualize the results of the correction using SCANet. We can observe that SCANet effectively identifies and corrects erroneous components in the assembly results produced by the MEPNet. Particularly in Figure~\ref{fig:9} (d), not only are there errors present in the current step, but multiple assembly errors are also in the already assembled parts (as 100\% correction cannot be guaranteed). However, in situations that have not been encountered during training, SCANet can still perform error correction, further demonstrating the robustness of SCANet.

\noindent \textbf{Ablation Experiments.} In the ablation study, we first adapted MEPNet by replacing its component encoder with our component pose encoder and added an assembly result input branch to make it suitable for the single-step assembly error correction task. We then compared it with our SCANet. 
Next, we removed the branch responsible for inputting assembly results from the backbone of SCANet to assess the necessity of the assembly-shape branch.
Then, we validated the role of the image encoder in the component pose encoder, as mentioned earlier, which can alleviate SCANet's problem of correcting components with the same shape but different 6D poses to the same pose.
Finally, we further removed the 6D pose encoder from the component pose encoder, retaining only the 3D voxel encoder. However, before encoding the 3D voxel data, we performed an RT matrix transformation on the voxel data in advance to verify the effectiveness of the 6D pose encoder.
As depicted in Table \ref{tab:table2}, although the improved version of MEPNet can correct misassembled results to some extent, its performance is significantly inferior to SCANet. Furthermore, we observed that after removing the 2D image encoder, while its correction rate (CR) is higher than that of the final SCANet, its misplacement rate (MPR) is also higher. As mentioned earlier, the absence of 2D image information causes the model to correct many components with the same shape but different poses to the same pose, thereby reducing the overall error correction capability. Finally, we observe that after removing the 6D pose encoder, the performance of SCANet drops sharply, highlighting the significant role played by this encoder.

\section{Conclusion and Discussion}

This paper introduces a novel task: `Single-Step Assembly Error Correction Task', which aims to address errors arising during component assembly. To support this task, we constructed the LEGO-ECA dataset based on Synthetic LEGO data. Additionally, we propose a new framework called SCANet, designed specifically for this task. Experimental results show that SCANet effectively identifies and corrects assembly errors, significantly enhancing assembly accuracy. \textbf{Future work}: Future efforts could focus on developing more advanced assembly error correction mechanisms for even better results. While our current approach targets the correction of errors in the present assembly step, the correction of errors from previous steps remains unexplored, representing a more challenging yet promising research direction. \textbf{Limitation}: A current limitation of our work is that the assembly correction task is only addressed in simulation. Whether this reflects real-world robotic scenarios remains to be seen, and extending this work to real-robot applications will be a focus of future research.

\section{Acknowledgment}
This work was supported by the National Youth Talent Support Program (8200800081), the National Natural Science Foundation of China (Grant Nos. 62376006 and 62136001). We gratefully acknowledge their support.








\nocite{*}

\bibliographystyle{IEEEtran}
\bibliography{main}


\end{document}